\documentclass[conference]{IEEEtran}
\IEEEoverridecommandlockouts
\usepackage{cite}
\usepackage{amsmath,amssymb,amsfonts}
\usepackage{algorithmic}
\usepackage{graphicx}
\usepackage{textcomp}
\usepackage{xcolor}
\usepackage[tableposition = top]{caption}
\usepackage[T1]{fontenc}
\usepackage{hyperref}
\usepackage{float}
\usepackage{afterpage}
\usepackage{mwe}
\usepackage{subcaption}
\usepackage[export]{adjustbox}

\def\BibTeX{{\rm B\kern-.05em{\sc i\kern-.025em b}\kern-.08em
    T\kern-.1667em\lower.7ex\hbox{E}\kern-.125emX}}
\begin{document}

\title{Multistream Graph Attention Networks for Wind Speed Forecasting\\}

\author{\IEEEauthorblockN{Dogan Aykas}
\IEEEauthorblockA{\textit{Department of Knowledge Engineering} \\
\textit{Maastricht University}\\
Maastricht, The Netherlands \\
d.aykas@student.maastrichtuniversity.nl}
\and
\IEEEauthorblockN{Siamak Mehrkanoon*\thanks{*corresponding author.}}
\IEEEauthorblockA{\textit{Department of Knowledge Engineering} \\
\textit{Maastricht University}\\
Maastricht, The Netherlands \\
siamak.mehrkanoon@maastrichtuniversity.nl}
}

\maketitle

\begin{abstract}
Reliable and accurate wind speed prediction has significant impact in many industrial sectors such as economic, business and management among others. This paper presents a new model for wind speed prediction based on Graph Attention Networks (GAT). In particular, the proposed model extends GAT architecture by equipping it with a learnable adjacency matrix as well as incorporating a new attention mechanism with the aim of obtaining attention scores per weather variable. The output of the GAT based model is combined with the LSTM layer in order to exploit both the spatial and temporal characteristics of the multivariate multidimensional historical weather data. Real weather data collected from several cities in Denmark and Netherlands are used to conduct the experiments and evaluate the performance of the proposed model. We show that in comparison to previous architectures used for wind speed prediction, the proposed model is able to better learn the complex input-output relationships of the weather data. Furthermore, thanks to the learned attention weights, the model provides an additional insights on the most important weather variables and cities for the studied prediction task.

\end{abstract}


\section{Introduction}
Wind speed forecasting has gained a lot of attention in recent years since wind speed plays a significant role in human life. Sectors like agriculture, construction and energy can greatly benefit from an accurate estimation of several time steps ahead wind speed values. The field of weather forecasting generally utilizes Numerical Weather Prediction (NWP) \cite{nwp} methods which are based on mathematical formulations using several atmospheric features. With this method, the predictions are made by using equations that are derived with the aid of laws of physics and thermodynamics. Although NWP is a powerful method to such forecasting tasks, it also has its drawbacks such as requiring too much computational power \cite{nwp_a}. Moreover, due to the assumptions of weather variables used in this method, forecasting with NWP might be sensitive to noise that may be present in the weather variable measurements \cite{siamak}. 

The availability of large training data allows data-driven models to learn the existing underlying complex pattern of the data. In addition, thanks to the advances in computational power, one is able to train deep machine learning models. Therefore, deep learning based models have recently been increasingly used in weather forecasting tasks as an alternative to NWP \cite{wind_nn, wind_p_f, wf2, wlstm, ismail2, jesus1, jesus2, tent, kevin2}. In particular, the literature of weather forecasting models has witnessed the use of Convolutional Neural Network based approaches \cite{siamak, kevin, ismail2, cnn_w}, Graph Convolutional Neural Networks \cite{gcn_t, gcn_2}, Recurrent Convolutional Neural Networks \cite{ismail1, convlstm} and Restricted Boltzmann Machines \cite{rbm, rbm2}.



This paper uses Graph Attention Networks (GAT) \cite{gat} as the backbone of the proposed model. Graph Attention Networks are a subclass of Graph Neural Networks designed for analyzing graph data. GAT learns the spatial relationship between different vertices of a graph and how the features of a vertex may affect features of other vertices. Here, we propose an extension of GAT architecture such that features of a vertex can be represented as a 2-dimensional matrix. The introduced model consists of GAT architecture followed by a Long Short Term Memory (LSTM) layer in order to utilize both the spatial and temporal aspect of the data, respectively. In contrast to vanilla GATs, which assume each vertex of a graph has one dimensional features, our proposed model includes GAT layers that compute the attention scores by allowing the vertices to have two dimensional features. This enhancement to GAT allows calculating attention of the vertices per weather variable, resulting in a more detailed and interpretable GAT structure. The authors in \cite{gcn_t} proposed an deep Graph Convolutional Networks model for wind speed prediction tasks. In particular, they treated the weather stations as nodes of a graph whose associated adjacency matrix is learnable. Here, following the lines of \cite{gcn_t}, the GAT layer in our proposed model is also equipped with learnable adjacency matrix. This helps the model to learn the connection between weather stations in an end-to-end fashion, This type of information is not typically available for such forecasting tasks.

In our experiments, two real life historical weather data from Dutch and Danish cities are used to predict the wind speed values in multiple weather stations for several hours ahead ranging from 2 up to 24 hours. The paper is organized as follows. A brief overview of the related work is given in section \ref{related_work}. The proposed model is described in section \ref{method}. The numerical experiments, discussion of the obtained results and comparison with other models are given in Section \ref{experiments}. The conclusion is drawn in section \ref{conclusion}.

\section{Related Work}
\label{related_work}
With the rise of impressive results obtained by data driven based models, the literature has witnessed the design of various deep learning architectures for weather forecasting tasks \cite{weather_nn, wf2, wf3, wf4,siamak,kevin}. The weather data can be considered as time series data where observations occur as a sequence. Recurrent neural networks have been often used for dealing with time series forecasting tasks \cite{weather_rnn}. Specifically, Long Short Term Memory (LSTM) models are preferred for such tasks due to their success in handling vanishing/exploding gradients \cite{lstm} which can possibly hinder the performance during training phase. However, vanilla LSTMs do not explicitly exploit the spatial features of data and thus, they are not able to capture the influences of the features observed in one city to the other cities and vice versa. 

On the other hand, Convolutional Neural Networks (CNN) are widely used, especially for computer vision tasks such as face recognition, image classification \cite{cnn_face}. While vanilla CNNs are able to process the spatial aspect of the data, it is unable to efficiently make use of the temporal characteristics which are important for such weather forecasting applications. The authors in \cite{siamak}, introduced a 3D-CNN based model for weather elements forecasting which exploit the spatio-temporal multivariate weather data for learning shared representations. In \cite{tcn}, Temporal Convolutional Networks (TCN), a type of CNN which uses one dimensional filters in order to focus on the temporal aspect of data, are used in combination with vanilla CNNs for weather forecasting tasks . CNNs have the ability to process the spatio-temporal aspect of the data with the addition of TCNs.

Due the nature of the fixed size filters, CNNs have the tendency to treat observations from multiple stations as regular or Euclidean data \cite{regular}. 
Considering the spatial dimension of our data which corresponds to different weather stations, it can be highlighted that the underlying graph structure cannot be efficiently expressed under the the regular data setting. Therefore, the captured spatial relationship between observations are restricted to the filter size in CNNs. In the context of weather data, the spatial aspect of historical observations from multiple weather stations can be expressed more efficiently as a graph, with stations as vertices and connections between stations as edges. 

Graph Neural Networks (GNN) \cite{gnn}, has recently gained popularity for analyzing graph data. A variation of GNN, i.e. Graph Convolutional Networks (GCNs) \cite{gcn}, combines the basics of GNNs by working with graph data rather than assuming the data to be Euclidean, while maintaining a convolutional architecture. Although most variants of GCNs mainly deal with node classification tasks which require the vertex and edge information of the graph data to be known, there has been GCN adaptations in \cite{gcn_l, gcn_t} with a learnable adjacency matrix. The learnable adjacency matrix, which is also used in our proposed model, is significant for data with spatial features that are not represented explicitly in a graph format with a known adjacency matrix. This on one hand, allows regular data to be represented with an adjacency matrix usually required in GNN implementations. On the other hand, the adjacency matrix is learned during the training phase, which is beneficial for the spatial information extraction. Attention mechanisms have recently gained a lot of popularity, thanks to the transformer architecture introduced in \cite{transformers}. The attention mechanism has also been incorporated in GNN frameworks. For instance, GAT \cite{gat} uses self attention mechanism similar to the one in transformer model. In particular, GAT uses a self attention mechanism in order to calculate the attention coefficient between vertices of a graph. Furthermore, there have been implementations of GATs used for spatio-temporal data, specifically for traffic flow forecasting \cite{gatlstm}.


In what follows, we present a novel model which consists of an extension of GATs architecture followed by LSTM layer for wind speed prediction tasks. The proposed modifications of GATs offer an improved performance for weather element forecasting and further interpretability using the visualization of the learned attention weights of cities per weather variable.

\section{Proposed Method}
\label{method}
The proposed model consists of two components, i.e. two stream GAT layers followed by LSTM layer, in order to extract the spatial and temporal information of weather data respectively (see Fig. \ref{fig2}). The spatial component of our proposed model is inspired by classical GAT architecture \cite{gat}. More precisely, our proposed two stream GAT model exploits the graph related spatial information, while the LSTM layer captures the temporal dependencies between weather variables observed at different timesteps. The weather data is first cast to a graph where the weather stations form the vertices of a graph and each vertex contains two dimensions, i.e. weather variables and timesteps. Here, weather variables includes wind speed, wind direction, temperature, dew point, air pressure and rain amount. Timestep dimension contains values from historical observations of these weather variables. In the subsequent subsections, we first extend GAT model by equipping it with a learnable adjacency matrix as well as incorporating 2-dimensional node features. The later results in obtaining attention scores per weather variable. The proposed model architecture will then be discussed.

\subsection{GAT with Learnable Adjacency Matrix}
Often a graph is represented by an adjacency matrix. However, in case that the graph structure is not known, one may learn the edge weights of the graph. Previous studies \cite{gcn_2}, \cite{gcn_3} and \cite{gcn_t} made the adjacency matrix learnable which has shown to improve the predictive performance of the model. In addition, it can potentially provides an insights into how the features of a particular node is influenced by the features of the other nodes.  

The classical GAT model \cite{gat}, assumes that the adjacency matrix of the graph is known. Here, in contrast to the model in \cite{gat}, we make the adjacency matrix learnable. Therefore, we let the network to learn the graph spatial connections between the nodes in an end-to-end fashion. This learned information is later used in GAT formulation for obtaining the attention scores. Similar to \cite{gcn_t}, the following operations are applied to the adjacency matrix during the learning phase:

\begin{equation}
\label{eqn:e1}
\left\{\begin{array}{l}
\hat{A}=A+I, \\
\hat{A}=\frac{\hat{A}-\hat{A}_{m i n}}{\hat{A}_{m a x}-\hat{A}_{\min }}. 
\end{array}\right.
\end{equation}
Here, the identity matrix is added to the adjacency matrix $A$ in order to consider the possible self-loop connection to each vertex. Subsequently, min-max normalization is applied to the updated adjacency matrix $\hat{A}$.

\begin{figure*}[t]
    \centering
    \includegraphics[width=\textwidth]{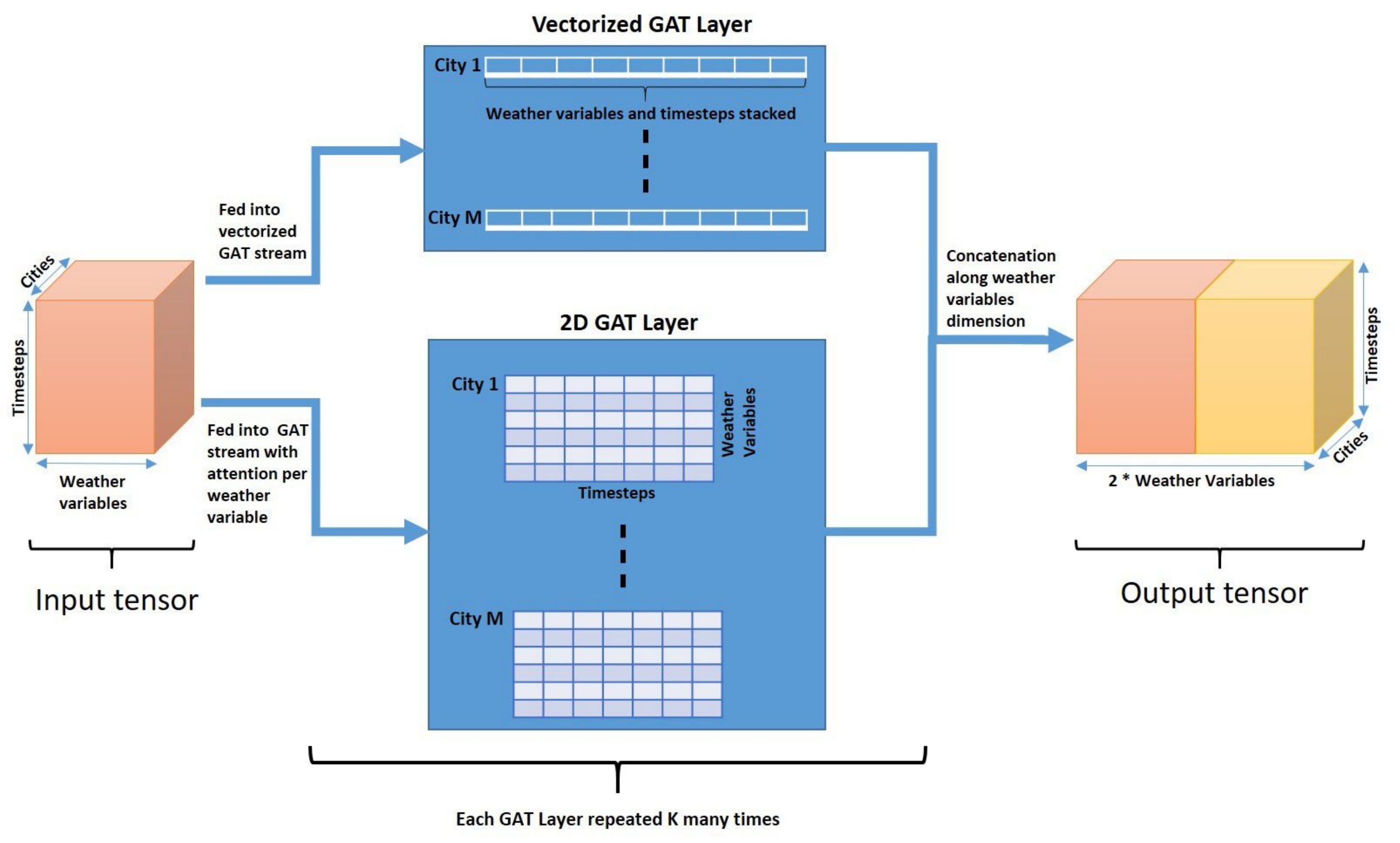}
    \caption{The two stream GAT layer used in our proposed model.}
    \label{fig1}
\end{figure*}

\subsection{2-Dimensional GAT}
In classical GAT formulation \cite{gat}, the feature values of vertices are considered to be one dimensional. Given the vertices and their one dimensional features, GAT computes attention scores which are then used for obtaining a new representation of each node in the graph. Following the lines of \cite{gat}, we assume that the set of node features are expressed as ${h}=\left\{{h}_{1}, {h}_{2}, \ldots, {h}_{N}\right\}$, where $N$ is the number of nodes. ${h}_{i} \in \mathbb{R}^{d}$, where $d$ is the number of features in each node \cite{gat}. The feature values are first linearly transformed to a pre-selected $d^{\prime}$ number of features by means of a shared weight matrix. Then using a self attention mechanism, attention scores of each vertex with respect to others are calculated and included in the transformation of feature values as a dot product. More precisely, given the $i$-th and $j$-th vertex pair, the attention score $\alpha_{i,j}$ is computed as follows \cite{gat}: 
\begin{equation}
\label{eqn:e2}
\alpha_{i j}=\frac{\exp \left(\text { LeakyReLU }\left({a}^{T}\left[{W} {h}_{i} \| {W} {h}_{j}\right]\right)\right)}{\sum_{k \in \mathcal{N}} \exp \left(\operatorname{LeakyReLU}\left({a}^{T}\left[{W} {h}_{i} \| {W} {h}_{k}\right]\right)\right)}.
\end{equation}
Equation (\ref{eqn:e2}) shows how features of the $i$-th and $j$-th vertex pair are transformed into $d^{\prime}$ dimensions and concatenated, i.e. $\left[{W} {h}_{i} \| {W} {h}_{j}\right] \in \mathbb{R}^{2d^{\prime}}$. Here, ${h}_{i} \in \mathbb{R}^{d}$, and ${h}_{j} \in \mathbb{R}^{d}$ refer to the feature vectors of the $i$-th and $j$-th vertex respectively. ${W} \in \mathbb{R}^{d^{\prime} \times d}$ is the shared weight matrix which linearly transforms features of each vertex. In GAT, the self attention mechanism is modeled by a single layer feed-forward neural network parameterized by ${a} \in \mathbb{R}^{2 d^{\prime}}$. The attention mechanism outputs a scalar. Next the LeakyRELU nonlinearity is applied with negative slope of 0.2. followed by a softmax to normalize the attention score. In GAT \cite{gat}, the softmax is applied with respect to all vertices within the neighborhood of the $j$-th vertex which is pre-determined by the adjacency matrix of the graph. However, in our case, adjacency matrix is learned during the training phase without imposing any constraint on the connectivity between vertices of the graph. Therefore, in our case all nodes are connected and the strength between the nodes are learnt through the elements of the adjacency matrix. The linearly transformed features of vertices are multiplied with the attention scores and then the sum is taken with respect to neighboring vertices \cite{gat}.

As previously stated, in our proposed model the nodes can have 2-dimensional features, i.e. weather variables and timesteps. Here, we extend GAT model so it can operate on 2-dimensional features and compute the attention scores per weather variable. To this end, we use Eq. (\ref{eqn:e2}), with a $W$ of size $\mathbb{R}^{t^{\prime} \times t}$ and ${h_{i}}$ of size $\mathbb{R}^{t \times d}$. Here, the weight matrix $W$ is used to transform the timestep dimension. The concatenation of ${W} {h}_{i}$ and ${W} {h}_{j}$, i.e. $\left[{W} {h}_{i} \| {W} {h}_{j}\right]$, will result in a matrix of size $2t^{\prime}\times d$. In contract to classical GAT formulation, here the product between the self attention coefficients ${a} \in \mathbb{R}^{2 t^{\prime}}$ and the concatenated nodes features   
$\left[{W} {h}_{i} \| {W} {h}_{j}\right]$ will result in a vector of size $d$, i.e. number of weather variables, instead of a scalar which is the case for GAT \cite{gat}. Next, the new representation of the $i$-th vertex is computed as follows:
\begin{equation}
\label{eqn:e3}
\hat{h}_{i}=\sigma\left(\sum_{j \in \mathcal{N}} {W} {h}_{j} \textrm{diag}({\alpha_{i j}^{1},..., \alpha_{i j}^{d}})\right),
\end{equation}
where, $\alpha_{i j}^{p}$ refers to the attention scores between $i$-th and $j$-th nodes for the $p$-th weather variable. Similar to \cite{gat}, we apply $K$ independent attention mechanisms and concatenate their results to obtain the final representation of the nodes as follows:

\begin{equation}
\label{eqn:e4}
\hat{h}_{i}=\|_{k=1}^{K}  \sigma\left(\sum_{j \in \mathcal{N}} {W}^{k} {h}_{j} \left[\textrm{diag}({\alpha_{i j}^{1},..., \alpha_{i j}^{d}})\right]^{k}\right).
\end{equation}
After obtaining $\hat{h}_i$, 
the final representation of the $i$-th node is computed as follows \cite{gcn_t}:
\begin{equation}
\label{eqn:e5}
\left\{\begin{array}{l}
\hat{D}_{i i}=\sum_{j} \hat{A}_{i j}, \\
\tilde{h}_{i}= \hat{h}_{i} \left(\hat{D}^{-\frac{1}{2}} \hat{A} \hat{D}^{-\frac{1}{2}} \right).
\end{array}\right.
\end{equation}
Here, diagonal matrix $D$ is calculated based on the obtained $\hat{A}$ in Eq. (\ref{eqn:e1}). Following this step, the symmetric normalization, i.e. $\hat{D}^{-\frac{1}{2}} \hat{A} \hat{D}^{-\frac{1}{2}}$, is applied . Finally, the features of a $i$-th node shown as $\hat{h}_{i}$ is multiplied with the resulting transformed matrix in order to calculate the new feature representation $\tilde{h}_{i}$. 
The explained GAT layers extracts spatial information of weather data. However, one can further enhance the model capability to capture temporal dependencies in the data and that how different observations at different timesteps affect one another. To this end, we use Long Short Term Memory networks \cite{lstm}. After the spatial layer consisting of two stream GAT layers, the output tensor is reshaped and fed into a LSTM network consisting of a single number of recurrent layer. 

\subsection{Proposed Architecture}
\begin{figure*}[t]
    \centering
    \includegraphics[width=\textwidth]{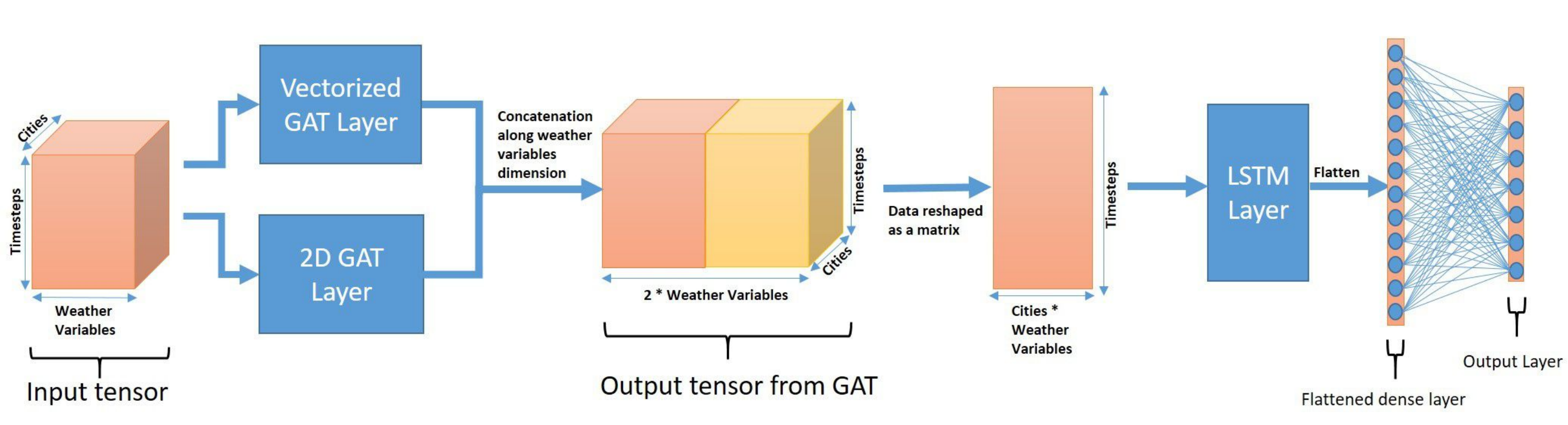}
    \caption{Overview of the proposed Multistream GAT model.}
    \label{fig2}
\end{figure*}

The proposed model consists of two streams of GAT layers in the spatial part, see Fig. \ref{fig1}. First stream functions as the default GAT formulation \cite{gat}, where each vertex has a one dimensional feature. Since the input data is originally 3 dimensional, i.e. weather variables $\times$ timestep $\times$ cities, the timestep and weather variables dimensions are stacked as a single dimensional vector when fed into this layer. The second stream is our proposed GAT architecture for obtaining attention per weather variable. In particular, in the second stream each vertex has two dimensional feature, therefore, the structural information of the weather data is kept unchanged. The output of each stream is a tensor after reshaping and in the end the output tensors to GATs are merged along feature dimension before being fed into the LSTM layer. The proposed model architecture is shown in Fig. \ref{fig2}.

\section{Numerical Experiments}
\label{experiments}
Two distinct weather datasets have been used for the evaluation of the proposed model: Netherlands and Denmark datasets. Both datasets contain hourly observations of several weather variables, collected simultaneously from different cities. Min-max normalization is applied to both dataset before the training phase. The inputs are first reshaped into a tensorial format which consist of 3 dimensions: weather variables, timesteps and cities. 

\subsection{Netherlands Dataset}
The Netherlands dataset contains hourly measurements of $6$ weather variables including wind speed, wind direction, temperature, dew point, air pressure and rain amount. For the Netherlands, the observations are collected from Schiphol, De Bilt, Leeuwarden, Eelde, Rotterdam, Eindhoven and Maastricht weather stations. Each training instance has $7\times 30\times 6$ shape, corresponding to weather variables $\times$ timesteps $\times$ cities. In our experiments, the number of timesteps is set to 30 as it is empirically found to be the optimal value. For this dataset, the target wind speed variable is predicted for 2, 4, 6, 8 and 10 hours ahead. The data contains observations from January 1, 2011 to March 29, 2020. The samples observed before January 2019 are used for training and validation, while samples from January 2019 to March 2020 are used for testing.

\afterpage{

\begin{table*}[t!]
\caption{The performance of the proposed model and other baseline models for the Netherlands dataset.}
\centering
  
\begin{tabular}{l|ccccc|ccccc}
\hline & \multicolumn{5}{|c|} { MAE } & \multicolumn{5}{|c} { MSE } \\
\hline Model & 2h ahead & 4h ahead & 6h ahead & 8h ahead & 10h ahead & 2h ahead & 4h ahead & 6h ahead & 8h ahead & 10h ahead \\
\hline 2D \cite{siamak} & $8.18$ & $10.08$ & $12.03$ & $13.15$ & $14.51$ & $118.35$ & $179.38$ & $253.60$ & $303.28$ & $369.16$ \\
2D + Attention \cite{siamak} & $8.10$ & $10.09$ & $11.83$ & $13.10$ & $14.13$ & $116.09$ & $180.69$ & $247.22$ & $300.41$ & $351.26$ \\
2D + Upscaling \cite{siamak} & $8.24$ & $10.22$ & $11.83$ & $13.74$ & $14.80$ & $120.65$ & $183.87$ & $248.85$ & $332.87$ & $387.92$ \\
3D \cite{siamak}& $8.05$ & $10.15$ & $11.93$ & $13.01$ & $14.24$ & $115.17$ & $183.55$ & $251.11$ & $294.39$ & $355.44$ \\
Multidimensional \cite{kevin} & $8.10$ & $10.03$ & $11.46$ & $12.79$ & $13.81$ & $115.92$ & $178.39$ & $228.97$ & $283.13$ & $336.29$ \\
WeatherGCNet \cite{gcn_t} & $7.96$ & $9.97$ & $11.16$ & $12.30$ & $13.33$ & $111.83$ & $174.55$ & $219.45$ & $265.71$ & $309.82$ \\
WeatherGCNet with $\gamma$ \cite{gcn_t} & $7.97$ & $9.74$ & $10.99$ & $12.44$ & $13.55$ & $113.32$ & $168.08$ & $212.30$ & $272.92$ & $319.05$ \\
Multistream GAT & $\underline{7.88}$ & $\underline{9.52}$ & $\underline{10.81}$ & $\underline{12.09}$ & $\underline{13.25}$ & $\underline{110.39}$ & $\underline{161.07}$ & $\underline{208.03}$ & $\underline{258.39}$ & $\underline{308.57}$ \\
\hline
\end{tabular}
\label{Tab:T1}

\end{table*}

\begin{table*}[t!]
\centering
\caption{The performance of the proposed model and other baseline models for the Denmark dataset.}
\begin{tabular}{l|cccc|cccc}
\hline & \multicolumn{4}{|c|} { MAE } & \multicolumn{4}{c} { MSE } \\
\hline Model & 6h ahead & 12h ahead & 18h ahead & 24h ahead & 6h ahead & 12h ahead & 18h ahead & 24h ahead \\
\hline 2D \cite{siamak} & $1.304$ & $1.746$ & $1.930$ & $2.004$ & $2.824$ & $5.088$ & $6.120$ & $6.610$ \\
2D + Attention \cite{siamak} & $1.313$ & $1.715$ & $1.905$ & $1.950$ & $2.885$ & $4.896$ & $5.933$ & $6.201$ \\
2D + Upscaling \cite{siamak} & $1.307$ & $1.723$ & $1.858$ & $1.985$ & $2.826$ & $4.931$ & $5.639$ & $6.474$ \\
3D \cite{siamak} & $1.311$ & $1.677$ & $1.908$ & $1.957$ & $2.855$ & $4.595$ & $5.958$ & $6.238$ \\
Multidimensional \cite{kevin} & $1.302$ & $1.706$ & $1.873$ & $1.925$ & $2.804$ & $4.779$ & $5.773$ & $6.066$ \\
WeatherGCNet \cite{gcn_t} & $1.279$ & $1.638$ & $1.777$ & $1.869$ & $2.698$ & $4.407$ & $5.148$ & $5.641$ \\
WeatherGCNet with $\gamma$ \cite{gcn_t}& $1.267$ & $1.616$ & $1.767$ & $1.853$ & $2.684$ & $4.285$ & $5.096$ & $5.566$ \\
Multistream GAT & $\underline{1.253}$ & $\underline{1.578}$ & $\underline{1.747}$ & $\underline{1.843}$ & $\underline{2.604}$ & $\underline{4.077}$ & $\underline{4.977}$ & $\underline{5.494}$ \\

\hline
\end{tabular}
\label{Tab:T2}
\end{table*}

\begin{figure*}[h!]

    \centering
    \begin{subfigure}[b]{0.36\textwidth} 
        \includegraphics[width=\linewidth]{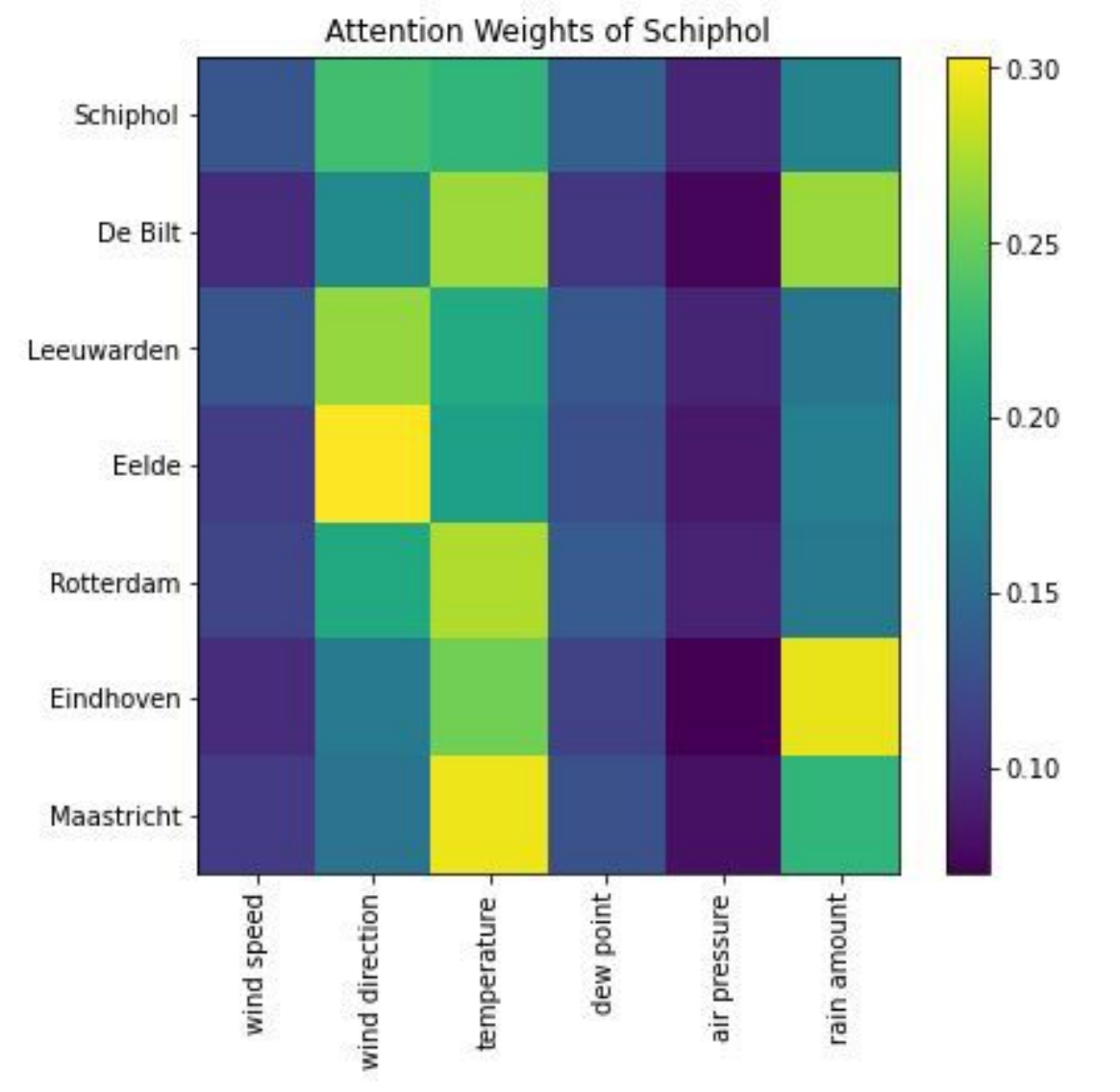}
        \caption{} 
    \end{subfigure}%
    \quad
    \begin{subfigure}[b]{0.36\textwidth} 
        \centering
        \includegraphics[width=\linewidth]{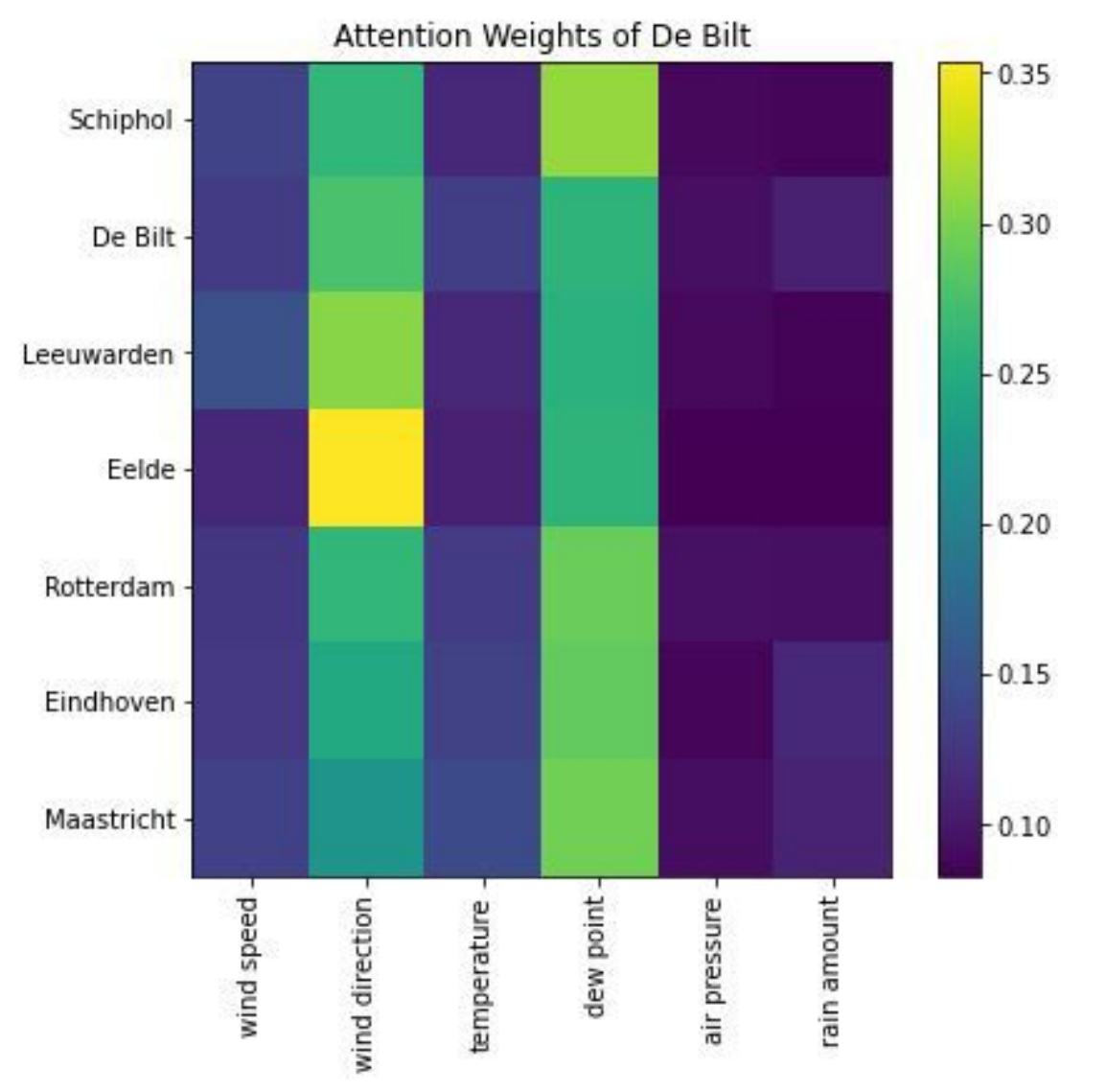}
        \caption{} 
    \end{subfigure}

    \begin{subfigure}[b]{0.36\textwidth} 
        \includegraphics[width=\linewidth]{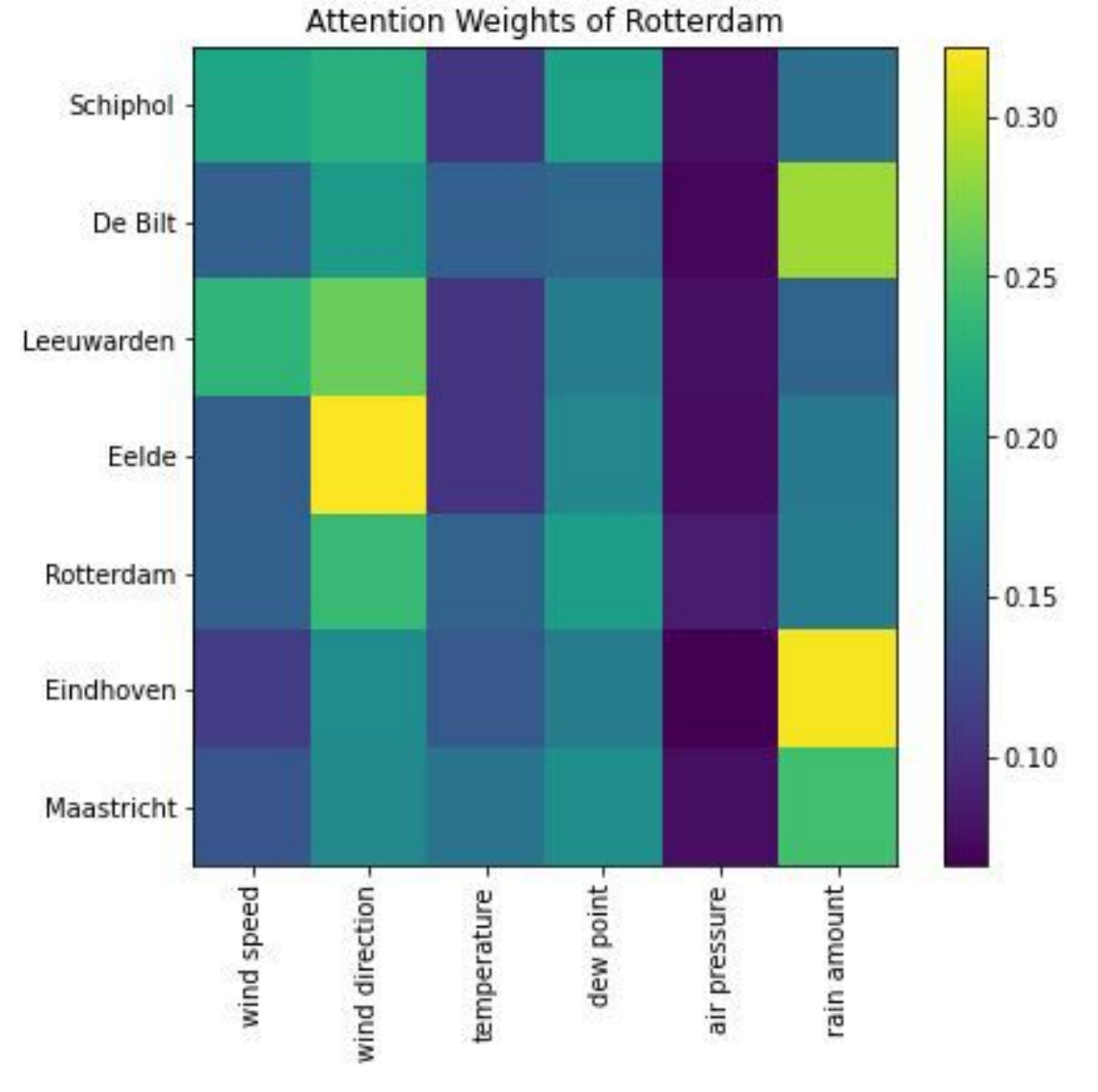}
        \caption{} 
    \end{subfigure}%
    \quad
    \begin{subfigure}[b]{0.36\textwidth} 
        \includegraphics[width=\linewidth]{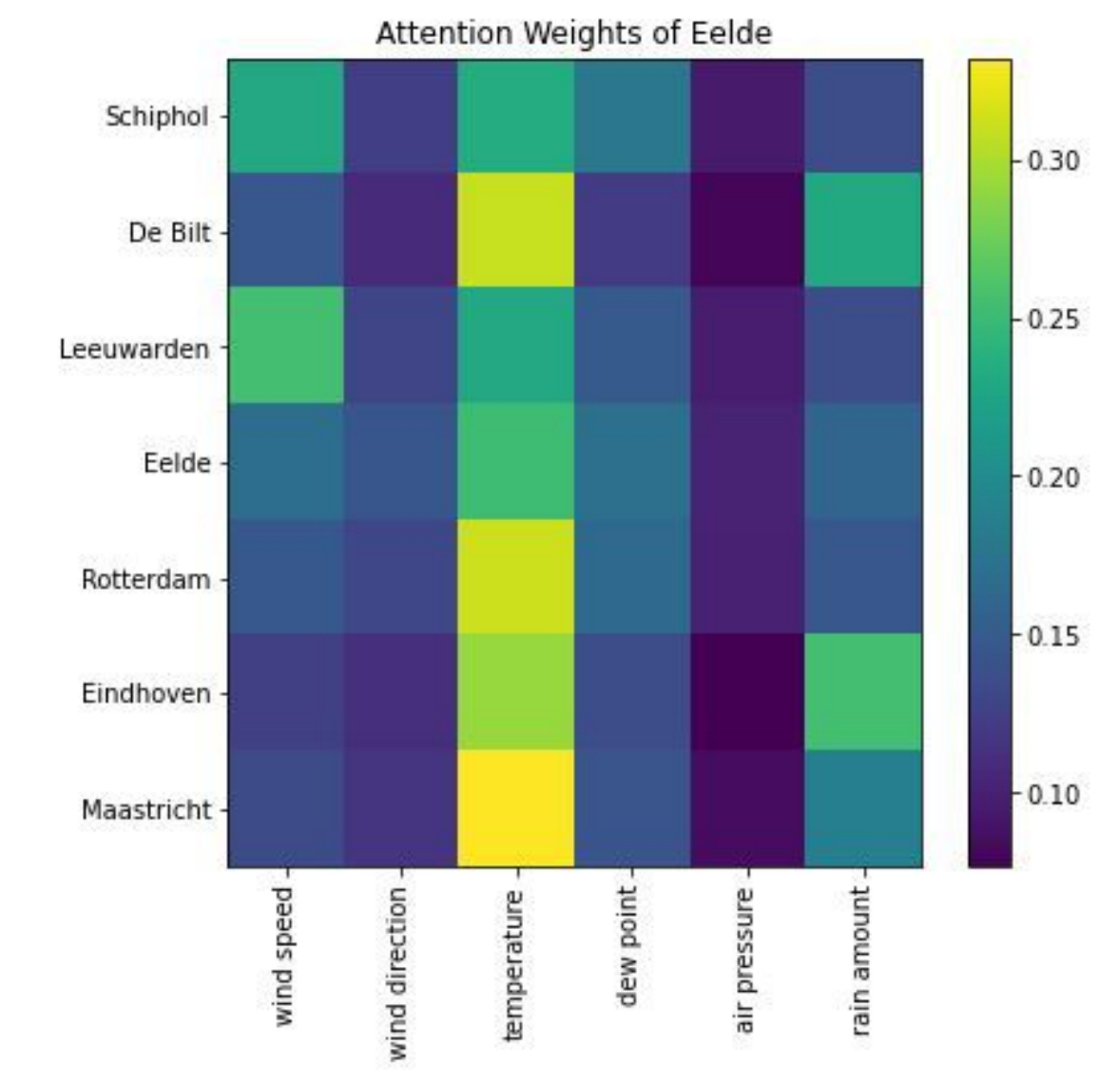}
        \caption{} 
    \end{subfigure}%
    \label{fig:foursubfig}

    \caption[The average and standard deviation of critical parameters ]
    {\small The attention weights of cities per weather variable.} 
    \label{fig3}
\end{figure*}
} 

\subsection{Denmark Dataset}
The Denmark data has the same 3 dimensions as the Netherlands dataset, with different sizes for weather variables and cities. Correspondingly, a total of four different weather variables i.e. temperature, pressure, wind speed and wind direction are observed from five different weather stations: Aalborg, Aarhus, Esbjerg, Odense and Roskilde. Each training instance has $5\times 30\times 4$ shape corresponding to weather variables $\times$ timesteps $\times$ cities. For this dataset, we aim at forecasting the wind speed of each city for 6, 12, 18 and 24 hours ahead as individual forecasting tasks. The measurement data from 2000 to 2010 is used. The samples before 2010 are used for training and validation, and the data in 2010 is used for testing.

\subsection{Results and Discussion}
The mean absolute error and mean squared error metrics are used to evaluate the performance of the tested models. The formulation of the used metrics are as follows: MAE$=\frac{\sum_{i=1}^{n}\left|y_{i}-\hat{y}_{i}\right|}{n}$ and MSE$=\frac{\sum_{i=1}^{n}\left(y_{i}-\hat{y}_{i}\right)^{2}}{n}$. 
Here, the real measurements and model predictions are denoted by $\hat y_{i}$ and $\hat y_{i}$ respectively and $n$ is the number of instances. The performance of the proposed model is compared with other models in \cite{gcn_t, kevin, siamak}. In order to have a fair comparison, the training/validation/test sets are identical for all of the models. It can be observed from Tables \ref{Tab:T1} and \ref{Tab:T2} that the proposed Multistream GAT model outperforms the other tested baseline models in terms of MAE and MSE scores for both Danish and Netherlands datasets at different timesteps. 
Furthermore, thanks to the learned attention coefficients per weather variable, the Multistream GAT model provides more detailed insights to the predictions. Although the used weather data has spatial characteristics, it does not necessarily contain underlying graph related edge information regarding how different cities are connected to each other. The hidden graph related knowledge is discovered and utilized, by means of the incorporation of learnable adjacency matrix in each of the GAT stream layer.

\begin{figure*}
    \centering
    
    \begin{subfigure}[t!]{0.48\textwidth}
    \centering
        
        \includegraphics[scale=0.75,valign=t]{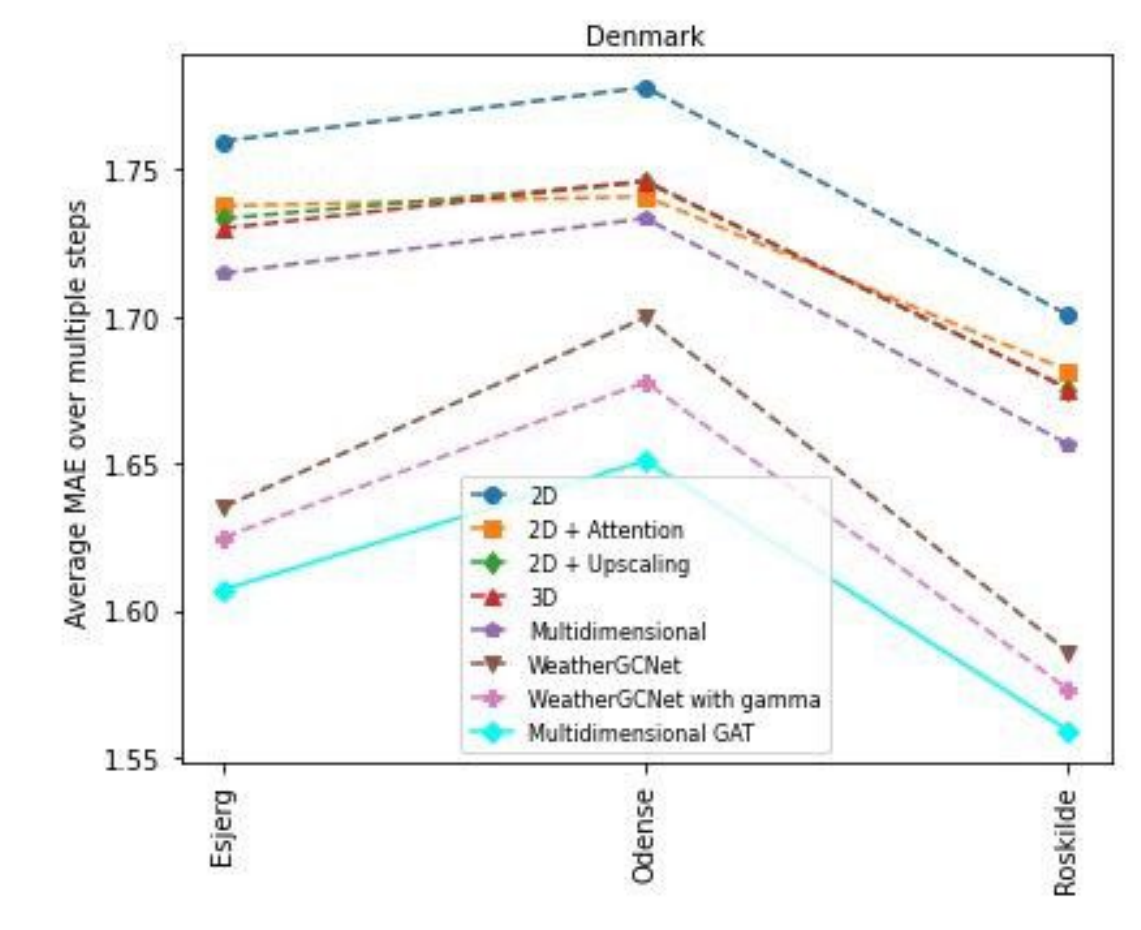}
        \caption[]%

    \end{subfigure}
    \hfill
    \begin{subfigure}[t!]{0.48\textwidth}  
        
        \centering
        
        \includegraphics[scale=0.71,valign=t]{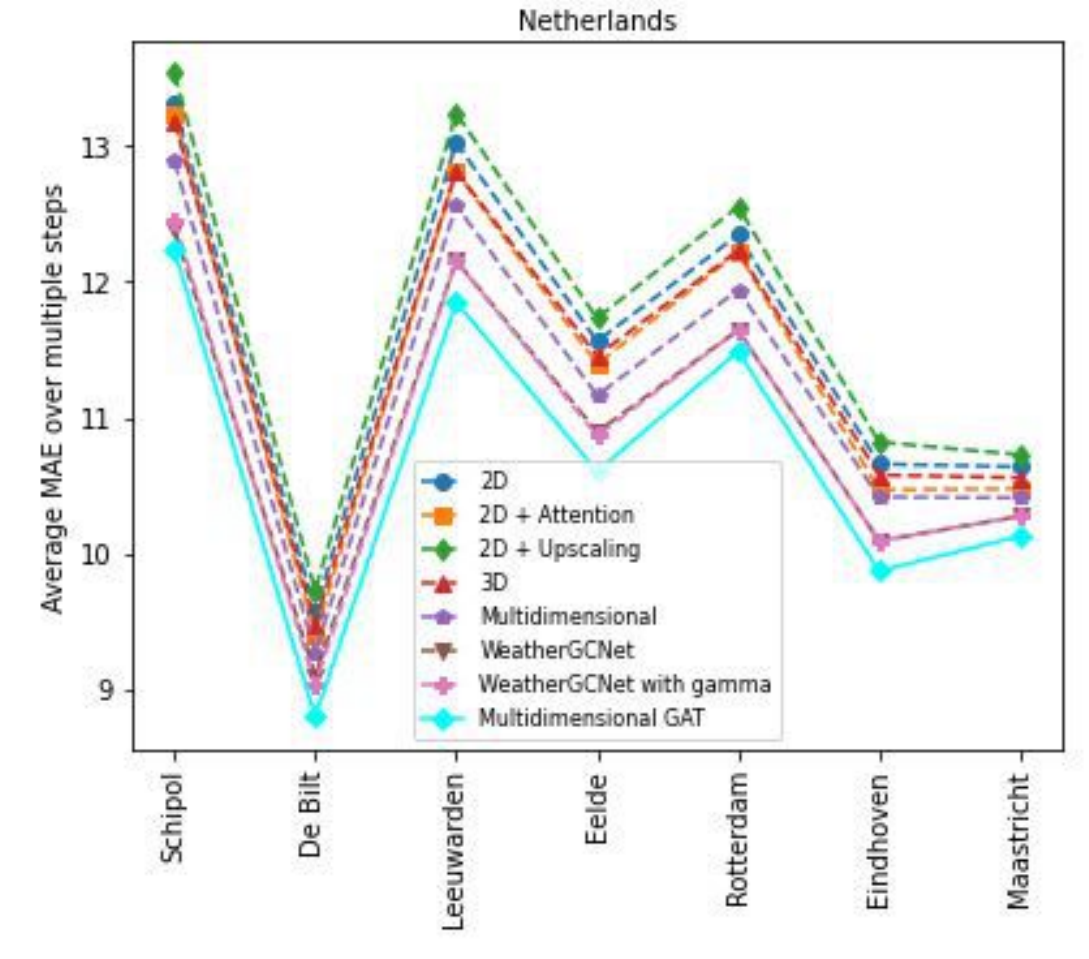}
        \caption[]%

    \end{subfigure}

    \caption[ The average and standard deviation of critical parameters ]
    {\small The average of the models performance across different time steps for each city. (a): Danish cities, (b): Dutch cities.} 
    \label{fig4}
\end{figure*}

\begin{figure*}
    \centering
    \begin{subfigure}[b]{0.475\textwidth}
        \centering
        \includegraphics[width=\textwidth]{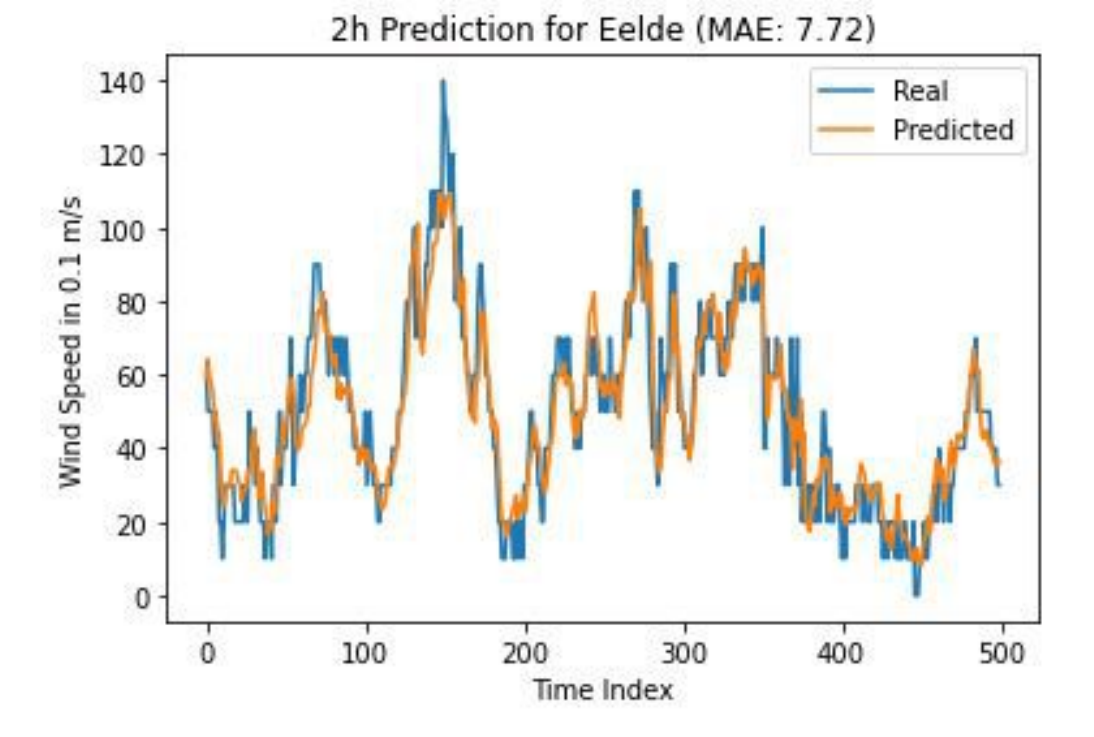}
        \caption[Network2]%
          
        \label{fig:mean and std of net14}
    \end{subfigure}
    \hfill
    \begin{subfigure}[b]{0.475\textwidth}  
        \centering 
        \includegraphics[width=\textwidth]{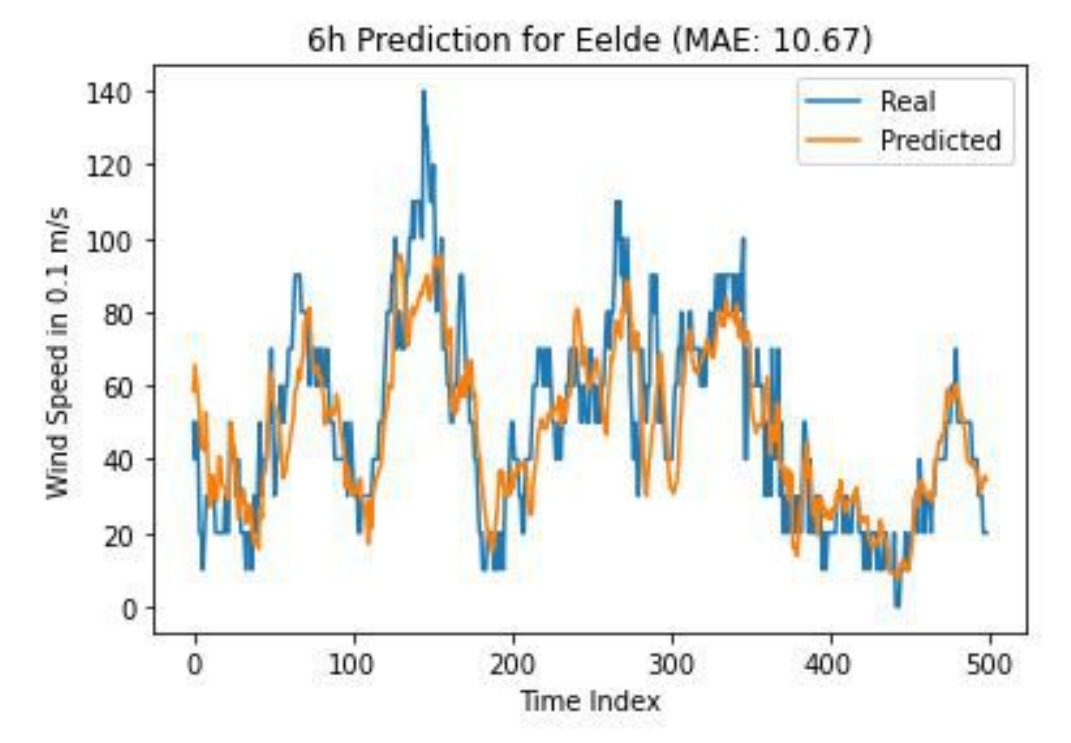}
        \caption[]%
          
        \label{fig:mean and std of net24}
    \end{subfigure}
    \vskip\baselineskip
    \begin{subfigure}[b]{0.475\textwidth}   
        \centering 
        \includegraphics[width=\textwidth]{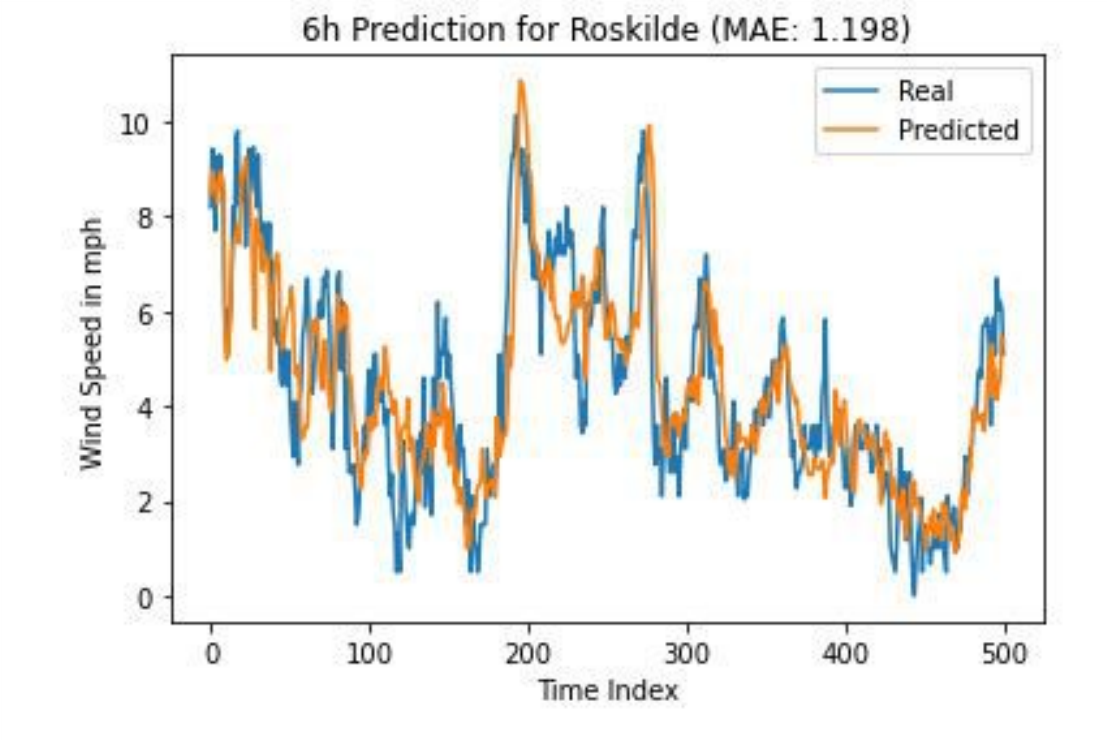}
        \caption[]%
          
        \label{fig:mean and std of net34}
    \end{subfigure}
    \hfill
    \begin{subfigure}[b]{0.475\textwidth}   
        \centering 
        \includegraphics[width=\textwidth]{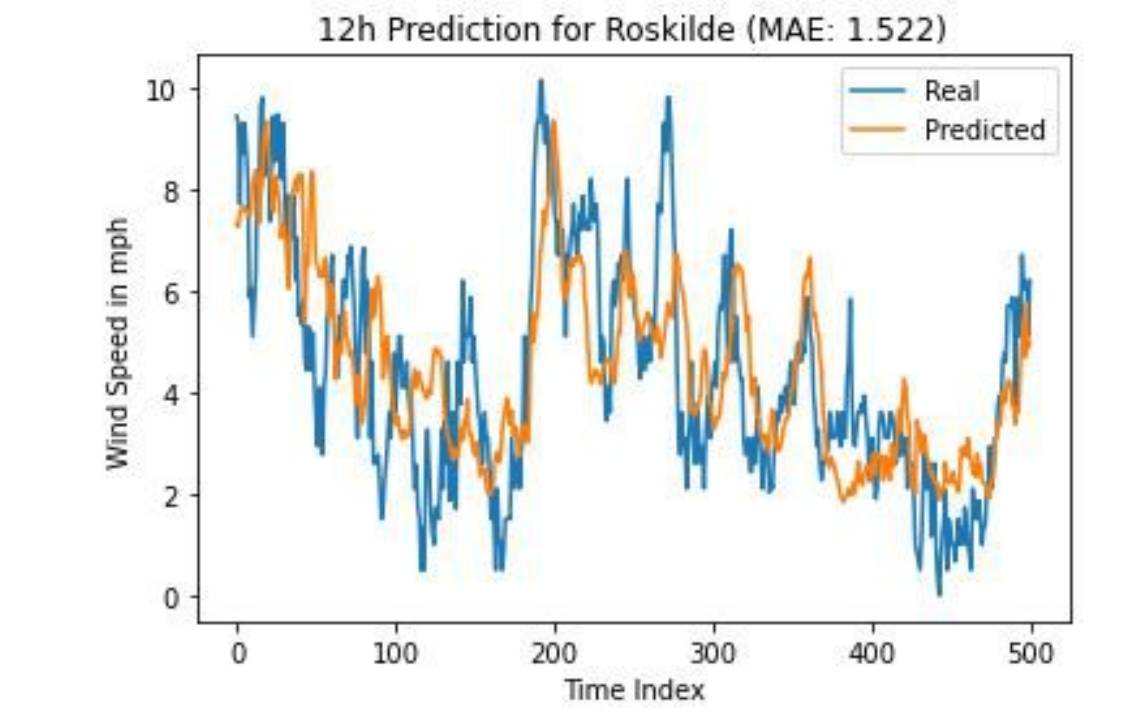}
        \caption[]%
           
        \label{fig:mean and std of net44}
    \end{subfigure}
    \caption[The average and standard deviation of critical parameters]
    {\small The wind speed predictions for Eelde and Roskilde. (a) and (b): 2 and 6 hours ahead prediction for Eelde. (c) and (d): 6 and 12 hours a head prediction for Roskilde.} 
    \label{fig5}
\end{figure*}

The classical GAT architecture offers some form of interpretability in terms of the attention weights between different nodes. Our proposed Multistream GAT, makes it possible to localize the attentions in a more detailed way for every weather variable. The representation of the attention weights of four Dutch cities can be observed in Fig. \ref{fig3}. The colormaps represent how much wind speed of the target city is affected by the weather variables of other cities. For instance, Fig. \ref{fig4} (c) shows that the wind direction of Eelde causes a significant influence on the wind speed of Rotterdam, while the wind speed of Rotterdam seems to be mostly affected by the rain amount of Eindhoven. 
Fig. \ref{fig4} compares the performance of different models by their average MAE scores per city. It can be observed that the proposed Multistream GAT model achieves the best result in each city. Moreover, the comparison of real and predicted wind speed values of Eelde and Roskilde cities for multiple timesteps ahead are shown in Fig. \ref{fig5}.

The proposed multistream GAT model extracts both the spatial and temporal information of the data, thanks to the combination of GAT and LSTM layers.  
In particular, the adjusted weather variables of the cities are fed into the LSTM network so that the temporal related information can be analyzed and taken into consideration by the model. 

It should be noted that, the graph convolutional architecture \cite{gcn_t}, which is shown to be successful for modeling the weather data, utilizes GCNs and temporal convolutions to deal with the spatio-temporal characteristic of the data. However, here the combination of GAT with LSTM outperforms the previous convolution based attempts, since the GAT allows more in depth analysis with the help of self attention mechanisms related to the spatial information. Furthermore, the temporal information is possibly utilized more efficiently with the LSTM layer than
the one dimensional filters present in Temporal Convolutional Networks.

\section{Conclusion}
\label{conclusion}
In this paper, a novel multistream Graph Attention Network is proposed for wind speed forecasting. We show that combining a graph based architecture with a recurrent model gives rise to efficient information extraction with respect to the spatio-temporal data. Thanks to the incorporation of the learnable adjacency matrix as well as attention per weather variable, the model can better exploit the underlying complex spatio-temporal characteristic of the weather data. Furthermore, the visualization of the learned attention weights per weather variable, provides a better insight into the understanding of the data. The applicability of the proposed model is shown on two real life historical weather data and it improved the wind speed forecasting in multiple weather stations. The implementation of our proposed model including the trained models are available on Github \url{https://github.com/doganaykas/Multistream-Graph-Attention-Networks}.


\bibliographystyle{IEEEtran}  
\bibliography{main} 

\vspace{12pt}

\end{document}